\documentclass[letterpaper, 10 pt, conference]{IROS2021/ieeeconf}  

\IEEEoverridecommandlockouts                              

\usepackage{epsfig}
\usepackage{amsmath}
\usepackage{amssymb}
\usepackage{mathrsfs}
\usepackage[dvipsnames]{xcolor}
\usepackage{cite}
\usepackage{caption}

\usepackage[noend]{algpseudocode}
\usepackage{algorithm}
\usepackage{graphicx}
\graphicspath{{images/}}
\usepackage{xspace}
\usepackage{multirow}
\usepackage{url}

\usepackage{array,booktabs}

\newcolumntype{C}[1]{>{\centering\let\newline\\\arraybackslash\hspace{0pt}}p{#1}}
\newcolumntype{L}[1]{>{\let\newline\\\arraybackslash\hspace{0pt}}p{#1}}

\newcommand{\alberto}[1]{\textcolor{blue}{[Alberto: #1]}}

\newcommand{\fgiuliari}[1]{\textcolor{magenta}{[FrancescoG: #1]}}

\newcommand{\methname}{POMP++\xspace}

\makeatletter
\DeclareRobustCommand\onedot{\futurelet\@let@token\@onedot}
\def\@onedot{\ifx\@let@token.\else.\null\fi\xspace}

\def\eg{\emph{e.g}\onedot} 
\def\ie{\emph{i.e}\onedot}

\makeatother

\title{\LARGE \bf
\methname: Pomcp-based Active Visual Search \\in unknown indoor environments
}

\author{Francesco~Giuliari$^{1,2}$,
        Alberto~Castellini$^{3}$,
        Riccardo~Berra$^{3}$,
        Alessio~{Del Bue}$^{2}$, \\
        Alessandro~Farinelli$^{3}$, 
        Marco~Cristani$^{3}$,
        Francesco~Setti$^{3}$ and 
        Yiming~Wang$^{2,4}$
\thanks{$^{1}$Department of Electrical, Electronics and Telecommunication Engineering, University of Genoa, Italy}
\thanks{$^{2}$Visual Geometry and Modelling (VGM) and Pattern Analysis and Computer Vision (PAVIS), Fondazione Istituto Italiano di Tecnologia, Italy.}
\thanks{$^{3}$Department of Computer Science, University of Verona, Italy.}
\thanks{$^{4}$Deep Visual Learning (DVL) Unit, Fondazione Bruno Kessler, Italy.}%
\thanks{}
\thanks{© 2021 IEEE.  Personal use of this material is permitted.  Permission from IEEE must be obtained for all other uses, in any current or future media, including reprinting/republishing this material for advertising or promotional purposes, creating new collective works, for resale or redistribution to servers or lists, or reuse of any copyrighted component of this work in other works.}
}

\begin{document}

\maketitle
\thispagestyle{empty}
\pagestyle{empty}

\begin{abstract}
In this paper, we focus on the problem of learning online an optimal policy for Active Visual Search (AVS) of objects in unknown indoor environments. 
We propose \methname, a planning strategy that introduces a novel formulation on top of the classic Partially Observable Monte Carlo Planning (POMCP) framework, to allow training-free online policy learning in unknown environments. We present a new belief reinvigoration strategy that enables the use of POMCP with a dynamically growing state space to address the online generation of the floor map.
We evaluate our method on two public benchmark datasets, AVD that is acquired by real robotic platforms and Habitat ObjectNav that is rendered from real 3D scene scans, achieving the best success rate with an improvement of $>$10$\%$ over the state-of-the-art methods.
\end{abstract}


\section{Introduction}
\label{sec:intro}
Finding a specific object in an indoor environment is a human activity that is trivial to define but complex to encode into an autonomous agent like a robot. This task leverages several basic human skills including self-localisation, visual search, object detection, along all with the necessary motor skills for reaching the object of interest, especially in highly cluttered and dynamic environments. 


\begin{figure}[!t]
    \centering
    \includegraphics[width=0.9\linewidth]{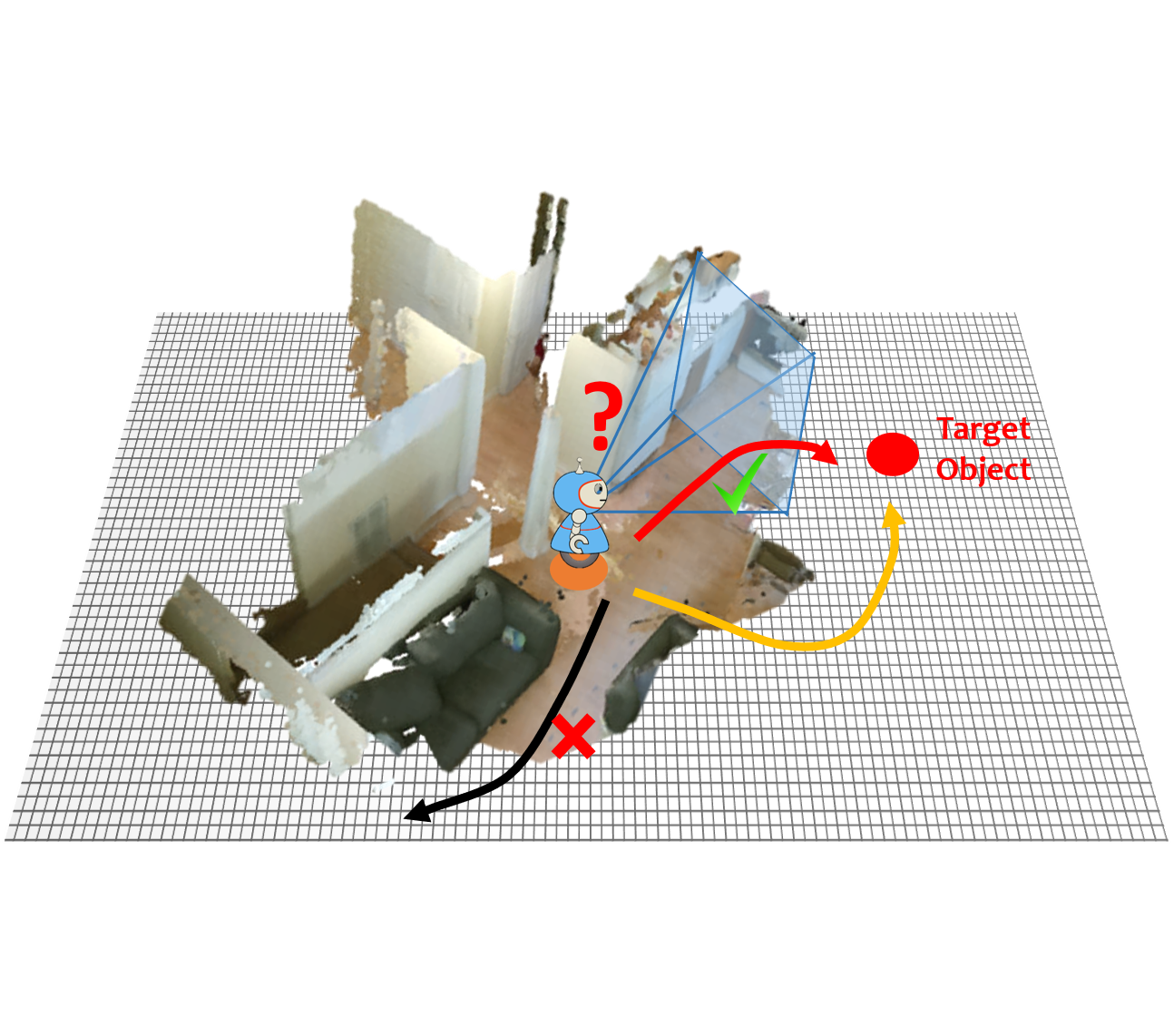}
    \vspace{-1cm}
    \caption{A robot is navigating in an \textit{unknown} environment with the task of visually searching for a target object (highlighted in red), \ie to have the object captured within the field of view of the camera. The robot is driven by a motion policy based on partially observed scene to visually detect the target object located in the unseen part of scene (highlighted in gray area) with the shortest travelled path (highlighted in red), to avoid longer trajectories (in yellow) or missing entirely the target (in black).} 
    \vspace{-0.2cm}
    \label{fig:intro_figure}
\end{figure}

In such context, this paper focuses on a specific task, Active Visual Search (AVS), where a robot is asked to navigate in indoor environments to search for a given object, with the performance being evaluated mainly on the success rate and the path efficiency (see Fig.~\ref{fig:intro_figure}). 
To address AVS, a joint solution for both visual perception and motion planning is needed.
Recent works mostly tackle this problem by intertwining deep Reinforcement Learning (RL) techniques, \eg deep recurrent Q-network (DRQN), with visual semantics, by either feeding deep visual embeddings to policy learning networks~\cite{schmid2019iros,ye2019ral} or obtaining 3D scene semantics to guide the planning~\cite{chaplot2020object}. 
These RL-based methods require a large amount of training data, \ie sequences of observations of various lengths, covering successful and unsuccessful episodes from real and simulated scenarios. Such pre-collected data may not be representative of the deployed (unknown) environments, thus possibly reducing the performance of a learning-based strategy. 

In this paper, we propose a method to specifically address AVS in \emph{unknown} environments without a priori knowledge of the area. While online policy learning has been used in previous approaches~\cite{wangpomp}, our method is the first to address this problem in the \emph{unknown} environment setting. 
Our approach, named \methname, is developed within the Partially Observable Monte Carlo Planning (POMCP) framework with two main adaptations introduced. We first present a formal definition of the AVS task in unknown environments following the Partially Observable Markov Decision Process (POMDP) formulation, in particular addressing the problem of a growing search space. 
%
Secondly, since the state space is dynamically changing in our problem, the original belief reinvigoration strategy of POMCP does not work as it requires a fixed set of states, we therefore propose a novel belief reinvigoration strategy that allows for an efficient path planning as the robot builds up the 2D map of the environment.
Once the POMCP module locates the target, the robot approaches it following the shortest path estimated by a local planner.


\begin{figure*}[t!]
  \centering
  \vspace{0.2cm}

  \includegraphics[width=0.9\textwidth,page=7,trim=40 30 50 15,clip]{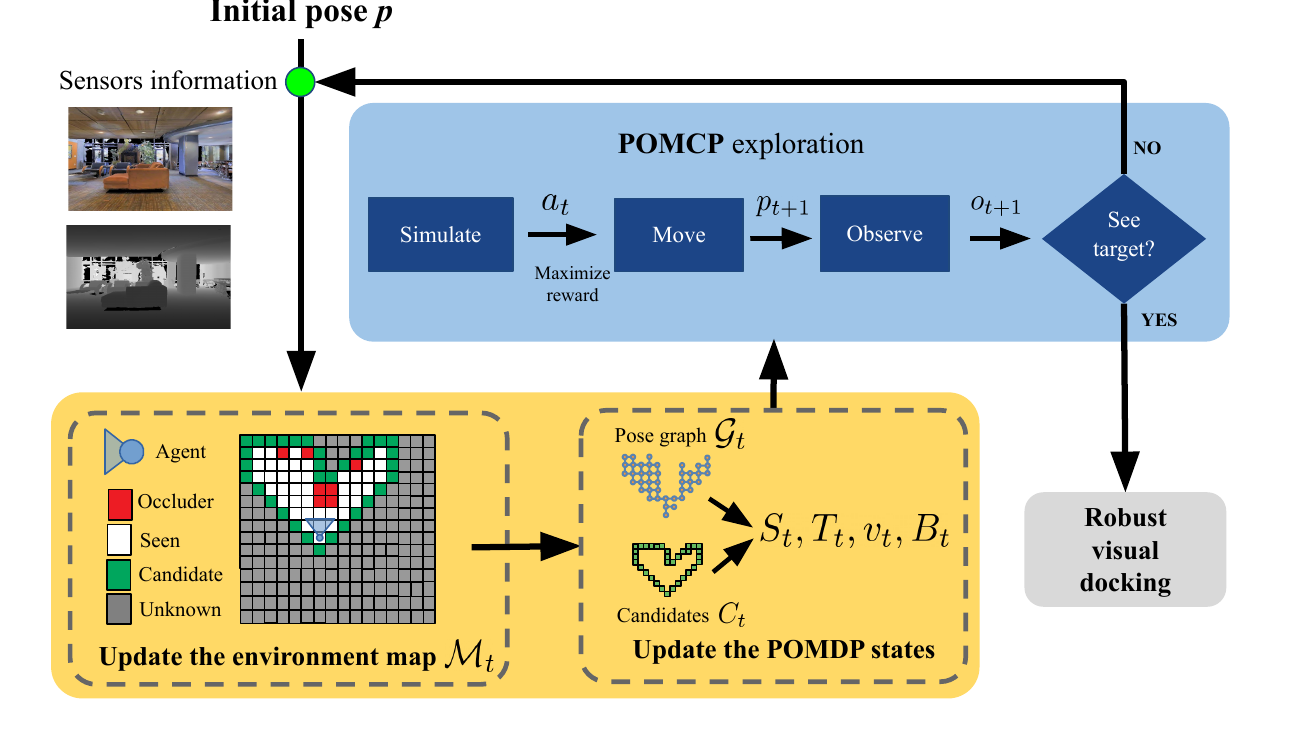}
  \vspace{-.5cm}
  \caption{Overall architecture of \methname. The yellow block represents the partial knowledge model of the environment that gets updated during the exploration, while the blue block represents the exploration strategy to locate the target. The mathematical notation is defined as: pose $p_t$, action $a_t$, observation $o_t$, POMDP state space $S_t$, POMDP transition model $T_t$, visibility function $v_t$, belief $B_t$.}
  \label{fig:method}
  \vspace{-0.4cm}
\end{figure*}

Our main technical contributions can be summarised as: 
\begin{enumerate}
    \item We formalise the AVS problem in unknown environments as a POMDP; 
    \item We integrate frontier-based exploration with POMCP planning for the AVS task, allowing the state space of the model to dynamically expand as the robot moves; 
    \item We propose a novel two-step belief reinvigoration strategy to address the dynamic state space within the POMCP framework.
\end{enumerate}
 
\methname is evaluated on public benchmark datasets including Active Vision Dataset (AVD) \cite{ammirato2017dataset} with sequences acquired by real robotic platforms and the Habitat ObjectNav Challenge \cite{batra2020objectnav} with observations rendered from real 3D scans. Our approach outperforms baselines and state-of-the-art (sota) approaches with a significant improvement of $>$10$\%$ in terms of the success rate, without performing any costly offline training.

\section{Related work}
\label{sec:soa}
We cover closely related works addressing the AVS task and the development and application of POMCP to exploration or vision domains.
\subsection{Active Visual Search}
\label{sec:soa:avs}
Earlier works addressing the AVS task often exploit target-specific inferences, such as object co-occurrences~\cite{Garvey1976,Wixson1994,Kollar2009,Sjoo2012,Aydemir2013}, while
recent works exploit Deep Reinforcement Learning techniques \cite{schmid2019iros,ye2019ral,chaplot2020object}, where visual neural embeddings are often used for the policy training. 
EAT~\cite{schmid2019iros} performs feature extraction from the current RGB image, and the image crop of the candidate target generated by a Region Proposal Network (RPN). The features are then fed into the action policy network. EAT is trained with twelve scans from the AVD \cite{ammirato2017dataset} dataset. 
Similarly, GAPLE~\cite{ye2019ral} uses deep visual features enriched by 3D information (from depth) for training the policy with a large dataset from 100 synthetic scenes.
%
Recent benchmarking efforts offer larger amount of data for training and foster more data-driven AVS solutions. For example, RoboTHOR~\cite{deitke2020robothor} provides observations rendered from photo-realistic rooms~\cite{deitke2020robothor}, while Habitat ObjectNav Challenge~\cite{batra2020objectnav} provides renderings from real 3D scans. SemExp~\cite{chaplot2020object}, the current best-performing method on the ObjectNav Challenge, takes RGB-D frames as input and learns a 3D scene semantics representation, which can then be used for selecting a long-term goal for reaching the target object. SemExp proved that the 3D scene semantics provide more efficient guidance compared to the runner-up solution driven by scene exploration.



In general, learning-based strategies leverage large amount of data to learn a model of the environment together with the motion policy, while online motion planning strategies have the advantage of being general and easy-to-deploy. A recent POMCP-based solution \cite{wangpomp} is able to achieve comparable search performance against sota data-driven methods without any offline training. However, this approach is only applicable to known environments, i.e. the 2D floor map of the scene is required. Instead, our method \methname follows the idea of online policy learning, and further proposes novel adaptations in terms of map representation and belief reinvigoration, which address unknown environments. 

%

\subsection{POMCP}
\label{sec:soa:pomcp}

\emph{Partially Observable Markov Decision Process (POMDP)} is an established framework for sequential planning under uncertainty~\cite{Kaelbling1998}. It allows to model dynamical processes in uncertain environments, where the uncertainty is related to actions and observations. 
While computing exact solutions for large POMDPs is computationally intractable~\cite{Papadimitriou1987}, impressive progress has been made by approximated \cite{Hauskrecht2000} and online \cite{Ross2008} solvers. One of the most used and efficient online solvers for POMDPs is Partially Observable Monte Carlo Planning (POMCP)~\cite{Silver2010} which combines a Monte Carlo update of the belief state with a \emph{Monte Carlo Tree Search (MCTS)} based policy~\cite{Coulom2006,Kocsis2006,Browne2012}, thus enabling the scalability to large state spaces. 
Most recent extensions of POMCP include applications to multi-agent problems~\cite{Amato2015}, reward maximisation with cost constraints~\cite{Lee2018} and the introduction of prior knowledge about state space to refine the belief space and increase policy performance~\cite{Castellini2019a}.

In this paper, we advance the state of the art by applying POMCP to AVS in unknown environments by proposing mechanisms to extend the state space, the transition and the observation models step by step. 
To address unknown environments, we exploit the elements of the frontier-based exploration theory~\cite{Yamauchi1997}.
Frontier-based exploration has been merged with POMDPs~\cite{Lauri2016}, but only for the exploration task. We instead focus on the AVS task and propose a novel POMDP formalisation and belief reinvigoration strategy to benefit effective path planning.



\section{Proposed Method}
\label{sec:method}
We consider the scenario where a robot moves in an \emph{unknown} environment, with the task of searching for a specific object. The robot explores the environment to: \emph{i)} observe the target object,
\emph{ii)} localise the object in the floor map, and \emph{iii)} finally approach the object, \ie moving close to the object location.
Each search episode is terminated once the robot believes the target object is sufficiently near to itself and visually detectable\footnote{The exact stop condition depends on the specific application. In the result section we consider standard evaluation protocols used in public benchmarks for AVS evaluation.}. 
We formulate the AVS problem as a POMDP and employ POMCP to compute the planning policy online. In the following section, we use the subscript $t$ to represent the elements at the current time and the subscript $t+1$ to represent the updated elements after a new sensor capturing and observation.

Our overall framework is shown in Fig.~\ref{fig:method}.
At each time step $t$ the robot locates at a pose $p_t=\{x_t,y_t,\theta_t\}$, where $x$ and $y$ are the coordinates on the 2D floor plane, and $\theta$ is the orientation. From that pose it receives a new capturing from a RGB-D sensor for updating the environment map $\mathcal{M}_t$, which encodes all the information about the environment that the robot has gathered until $t$. The environment map is used to: 1) create a pose graph $\mathcal{G}_t$, that includes all the reachable poses of the robot, 2) extract the candidate locations of the target object $\mathcal{C}_t$, namely the frontier positions bordering the explored and unknown cells, 3) update the internal states of the POMDP, and reinvigorate the belief to make it cover newly discovered candidate locations. 
Using the planning policy calculated by the POMCP exploration module, the robot reaches a new pose $p_{t+1}$ by taking an action $a_t$.
The process repeats until the target object is detected. Once it locates the object, the robot will approach it following the shortest path between the pose of the robot and the estimated position of the object, on the graph $\mathcal{G}_t$\footnote{We apply the same approaching strategy, i.e. robust visual docking, as in~\cite{wangpomp} that allows path re-planning with continuous observations to be more robust to poor object detection.}.
%
%




{\color{magenta}






}


In section~\ref{sec:method:pomdp} we present our new POMDP formulation that addresses the problem of exploring a search space that is initially unknown,
and grows as the robot discovers new parts of the environment. 
In section~\ref{sec:method:ourMethod} we detail how we use partial information about the environment to define the search space in POMCP to extract the exploration policy. We also show how the map of the scene can be updated every time when the robot discovers new parts of the environment.
Finally, section~\ref{sec:method:belief_resampling} focuses on a new belief reinvigoration strategy for updating the belief when new states are discovered. This is a key contribution of our approach since the state space in the standard POMCP is fixed and only the probability over states is updated step by step.

\subsection{POMDP formulation for Active Visual Search}
\label{sec:method:pomdp}
The POMDP used in our context is a tuple $(S,A,O,T,Z,R,\gamma)$ where
\begin{itemize}
\item $S$ is a finite set of partially observable \textit{states} representing combinations of robot poses and target object locations, whose cardinality is unknown as the environment is unknown; 
\item $A$ is the finite set of robot \textit{actions};
\item $O$:~$S\times A \rightarrow \Pi(Z)$ is the \textit{observation model};
\item $T$:~$S\times A \rightarrow \Pi(S)$ is the \textit{state-transition model} (with $\Pi(S)$ probability distribution over states in $S$); 
\item $Z$ is a finite set of \textit{observations}, namely, 1 if the searched object has been observed, 0 otherwise;
\item $R$:~$S \times A \rightarrow R$ is the \textit{reward function}, which is positive when the object is detected, and a negative when the robot moves in position where the object is not detected;
\item $\gamma \in [0,1)$ is a \textit{discount factor}.
\end{itemize}

The goal is to maximise the expected total discounted reward 
$\mathbb{E} \left[\sum_{t=0}^{\infty} \gamma^t R(s_t,a_t)\right]$, by choosing the best action $a_t$ in each state $s_t$. 
The term $\gamma$ reduces the impact of future rewards and ensures that the (infinite) sum convergences to a finite value. The partial observability of the state is dealt with by considering at each time step a probability distribution over all the states, called \emph{Belief} $B\in \mathcal{B}$. 
POMDP \emph{solvers} compute, in an exact or approximate way, a \emph{policy}, \ie a function $\pi$:~$\mathcal{B} \rightarrow A$ that maps beliefs to actions. We use POMCP~\cite{Silver2010} to compute the policy online. It is a recent algorithm employing Monte-Carlo Tree Search (MCTS) to solve POMDP. Every time an action has to be selected by the robot, POMCP performs controlled simulations using the known transition and observation models, and from those simulations it learns the approximate value of each action. Then it selects the action with the highest value. The algorithm uses a particle filter to represent the (approximated) belief and MCTS to drive the simulation process in an optimal way.

\subsection{Adaptation to unknown environments} 
\label{sec:method:ourMethod}

In unknown environments the state space $S$ can change dynamically because all the possible robot poses and object locations are not known \emph{a priori}, but instead discovered over time as the robot explores the environment. Here we describe how the main elements in the state space $S$ and, consequently, the transition model $T$, the observation model $O$, and the belief $B$ are updated at each time step. 



The pose graph $\mathcal{G}$ is composed of nodes representing possible robot poses and edges connecting poses reachable by the robot with a single action. 
We use $V^{\mathcal{G}}$ and $E^{\mathcal{G}}$ for the set of nodes and edges in $\mathcal{G}$ respectively. 
The pose graph $\mathcal{G}$ initially contains only poses in the field of view of the robot, while new nodes 
and edges 
are added as the exploration proceeds.
The calculation of the reachable poses from the current camera view is solved by scene reconstruction methods with the depth maps~\cite{3dreconstrucion}.

\begin{figure}[t]
\vspace{0.15cm}
\captionof{algorithm}{Map update after each step. We omit the subscript $t$ or $t+1$ for simplicity.}
\label{alg:map-update}
\begin{algorithmic}
\\\hrulefill
\Procedure{MapUpdate}{map $\mathcal{M}$, pose $p$}
\State retrieve image from RGB-D sensor from pose $p$
\State $rec\gets$ scene reconstruction from RGBD Image

\ForAll{cells $c\in \mathcal{M}$}
\If{$c$ in camera frustum}
\If{$c$ is unoccupied in $rec$}
    \State$c \gets seen$
\Else 
    \State$c \gets obstacle$
\EndIf
\EndIf
\EndFor
\ForAll{cells $c\in \mathcal{M}$}
\If{$c$ not camera frustum}
\If{$c$ is unoccupied in $rec$}
\If{$c$ is adjacent to $seen$ cell}
\State$c\gets candidate$
\Else
\State$c\gets unknown$
\EndIf
\EndIf
\EndIf
\EndFor
\State\Return $\mathcal{M}$
\EndProcedure
\\\hrulefill
\end{algorithmic}
\vspace{-0.4cm}
\end{figure}

We represent the \emph{environment map} $\mathcal{M}$ as a grid that covers the whole environment containing observed and unseen parts of the scene. 
%
We assign each cell of the map $\mathcal{M}$ with one of the four possible values: \emph{i)} \emph{occluder}, the cell in the real environment contains any occluders, e.g. walls, preventing the robot to see what is behind the cell;
\emph{ii)} \emph{seen}, 
a location which could contain the target, but has been observed and found empty by the robot;
\emph{iii)} \emph{candidate}, the cell is unknown but adjacent to a \emph{seen} cell, \ie a frontier cell;
\emph{iv)} \emph{unknown}, the cell is unseen and is not a candidate cell.
In the beginning, all the cells outside the field of view are initialised as \emph{unknown}. Cell values are then updated step by step as the robot explores the environment (see Algorithm~\ref{alg:map-update}). 

With the updated environment map $\mathcal{M}_{t+1}$, we can update the pose graph $\mathcal{G}_{t+1}$. For each node $p_i\in\mathcal{G}_t$ 
, if taking action $a_t$ leads the robot to a 
new pose $p_{j}$ in the environment which falls into a \emph{seen} cell of the updated map $\mathcal{M}_{t+1}$, then a node for pose $p_j$ is added to $\mathcal{G}_{t+1}$ with an edge $(p_i,p_j)$. 
We update the visibility function at each time step $t$ as $v_t: V^{\mathcal{G}}_t \times C_t \rightarrow \{ 0,1 \}$, where $V^{\mathcal{G}}_t$ is the set of nodes in $\mathcal{G}$ at time $t$ and $C_t$ is the set of \emph{candidate} cells, i.e. candidate object locations at time $t$.
The function takes a robot pose $p \in V_{\mathcal{G}}^t$, and a candidate object location $c \in C_t$, and returns 1 if $c$ is visible from $p$ (\ie no obstacle lying between $p$ and $c$ in the current map $\mathcal{M}_{t}$), 0 otherwise. Notice that, the aim of function $v_t$ is to address pose-cell visibility for all robot poses in $V^{\mathcal{G}}_t$ and all \emph{candidate} cells.

\begin{figure}[t]
\vspace{0.15cm}

\captionof{algorithm}{POMCP exploration. We omit subscripts $t$ and $t+1$ for simplicity.}
\label{alg:pomcp-exp}
\begin{algorithmic}
\\\hrulefill
\Procedure{POMCP-Exp}{initial pose $p$}
\State $\mathcal{M} \gets \text{init all cells to \textit{unknown}}$
\State $V^{\mathcal{G}} \gets \{p\}$
\State $E^{\mathcal{G}} \gets \emptyset$

\Repeat
\State $\mathcal{M}\gets\Call{MapUpdate}{p}$
\State $V^{\mathcal{G}} \gets V^{\mathcal{G}} \cup $  \{poses related to \emph{seen} cells in $\mathcal{M}$\}
\State $E^{\mathcal{G}} \gets$ edges between $p_i,p_j\in V^{\mathcal{G}}$
\State  $C\gets$ \{candidate cells in $\mathcal{M}$\}
\State $S \gets $ states considering updated $V^{\mathcal{G}} \text{and } C$
\State $T \gets \text{transition model from } E^{\mathcal{G}}$
 
\State $O \gets \text{visibility function on } \mathcal{M}$ 
\State $B \gets $ \Call{Belief reinvigoration}{$S$}


\State $a\gets \Call{POMCP}{S, T, O, p, B}$
\State $p \gets \Call{MOVE}{a}$

\State $o \gets $ \Call{Detector}{} \Comment{true if object is detected}
\Until $ o=$ True 
\EndProcedure
\\\hrulefill
\end{algorithmic}
\vspace{-0.4cm}
\end{figure}

\methname{} uses the elements described above to compute the optimal policy. Poses in $V^{\mathcal{G}}_t$ along with the \emph{candidate} cells in $\mathcal{M}_t$ are used to build the state set $S_t$; \emph{edges} in $E^{\mathcal{G}}_t$ are used in the transition model $T_t$, and the visibility function $v_t$ is used by the observation model $O_t$ in the simulation phase. Each state in $S_t$ represents the target object in a specific candidate cell. Algorithm~\ref{alg:pomcp-exp} describes how these states are updated during the exploration.
At each step $t$, POMCP executes $\beta$ number of simulations
to compute the optimal action to perform in the real environment. Each simulation uses a specific state, which assumes the object is in a specific {\emph candidate} cell. During the simulation the environment is not updated as there is no observation in the real environment.
After each step the observation model $O_t$ predicts if the robot sees the target object from the current pose and a reward is assigned to the action $a_t$.
The \emph{Reward} for $a_t$ is a high positive value if the robots moves in a position where it can detect the object, otherwise a small negative value is used to penalise longer paths.
Once the optimal action is identified by the simulations, the robot moves in the real environment and observes the object using an object detector. If the object is detected, the POMCP exploration terminates and the robot approaches the object with the robust visual docking strategy as in~\cite{wangpomp}.
Otherwise the environment map $\mathcal{M}_t$ and pose graph $\mathcal{G}_t$ are updated, and POMCP iterates until the maximum number of allowed steps $\mu$ is reached.

\subsection{Belief reinvigoration for dynamically growing state space}
\label{sec:method:belief_resampling}

\begin{figure}[t]
  \centering
  \includegraphics[width=1\linewidth,trim=0cm 0cm 7cm 0cm,clip=true]{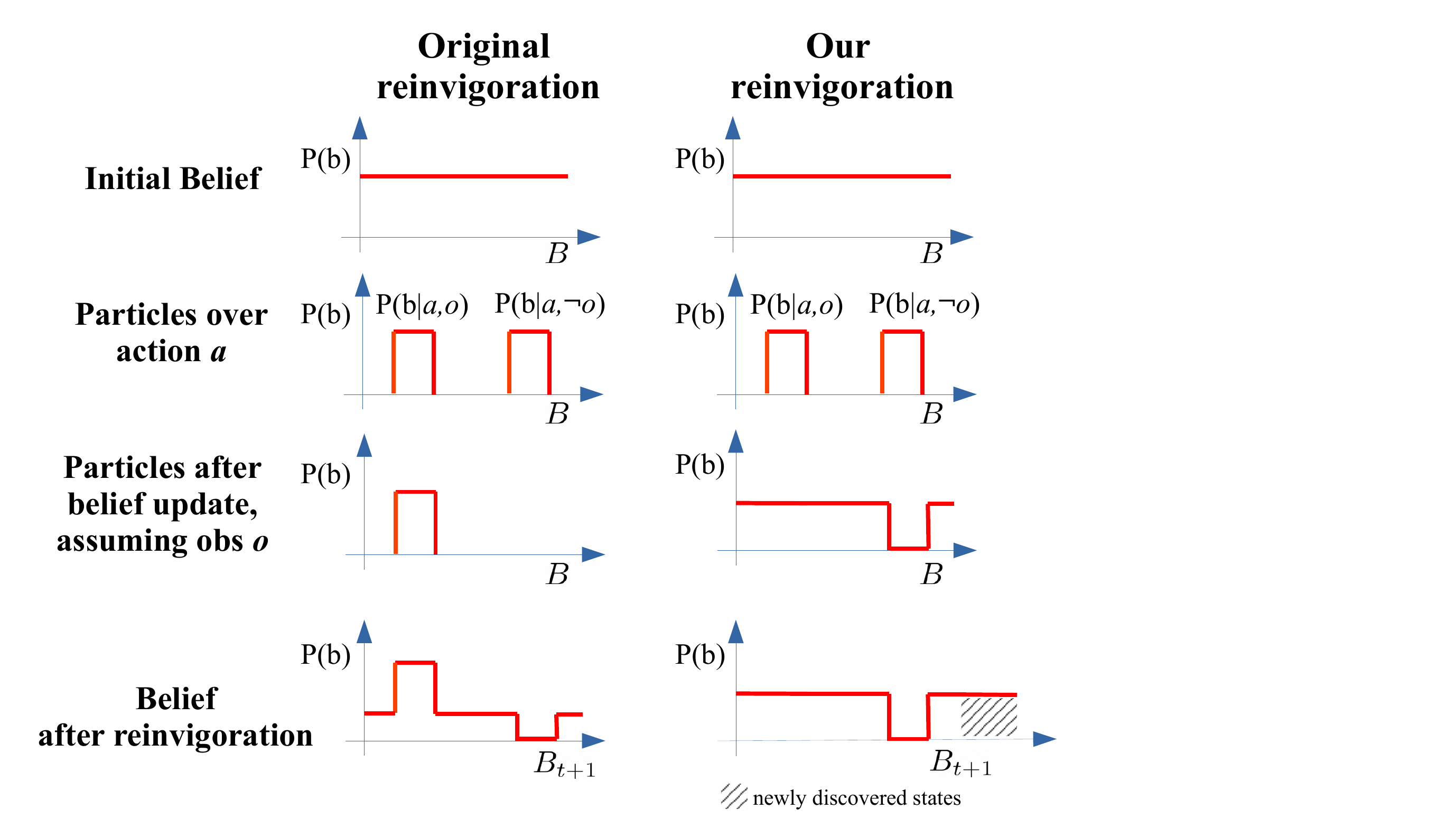}
  \caption{Difference between the original belief reinvigoration and our proposed reinvigoration strategy. The original reinvigoration penalises particles $b\in B$ that are not sampled by simulations taking $a$ as the first action, and receiving the observation $o$. Our new strategy, penalise only particles that were sampled by simulations where the simulated observation after the simulated action $a$ was different from the real world observation $o$. Our strategy also add the newly discovered candidate cells to the updated belief $B_{t+1}$.
  %
  }
  \label{fig:resampling}
\end{figure}


POMCP assigns the states to simulations by sampling particles $b$ from Belief $B$, which is updated over time when the robot makes an action and perform an observation in the real environment.

In standard POMCP, the belief update is performed considering only particles of the previous belief that are sampled by simulations whose first observation is the one observed in the real environment after performing the optimal action.
After the belief update, to avoid particle deprivation, new particles are added to the updated belief $B_{t+1}$ by sampling from the previous $B_t$ after executing the chosen action $a_t$. In such way, belief $B_{t+1}$ tends to have a much higher probability in states that were already present before the reinvigoration, thus discouraging actions towards newly added states, as can be seen in Figure~\ref{fig:resampling}.


Moreover, when dealing with an unknown environment where the state space changes over time, the particles at time $t$ do not account for the part of the environment that will be observed at time $t+1$. Hence the states in the particle filter, after the traditional belief update, cannot represent correctly the state space at time $t+1$. 
To address the above-mentioned limitations of the belief update strategy in standard POMCP, we propose a two-step belief reinvigoration strategy. 
First, we map all the states in $B_t$ to $B_{t+1}$, and remove the particles in simulations with their simulated observation different from the one made by the robot in the real environment.
In this way, states in $B_t$ are not penalised in $B_{t+1}$ if they are not used as initial states by the simulations.
Second, we add a new set of particles to $B_{t+1}$ sampled with a uniform distribution over the new candidate cells introduced at time $t+1$. Note that, without adding these states to $B_{t+1}$, simulations in next steps will assume that the object could never be in one of these newly discovered parts of the environment.
The proposed strategy allows to tune the percentage of new particles introduced in the second step with respect to the previously discovered candidate cells. By doing so, we can tune the attention the robot puts on the newly discovered frontier cells. 

\section{Experiments}
\label{sec:exp}

\begin{figure}[t!]
\begin{center}
	\begin{tabular}{@{}c@{}c}
		\includegraphics[width=0.23\textwidth]{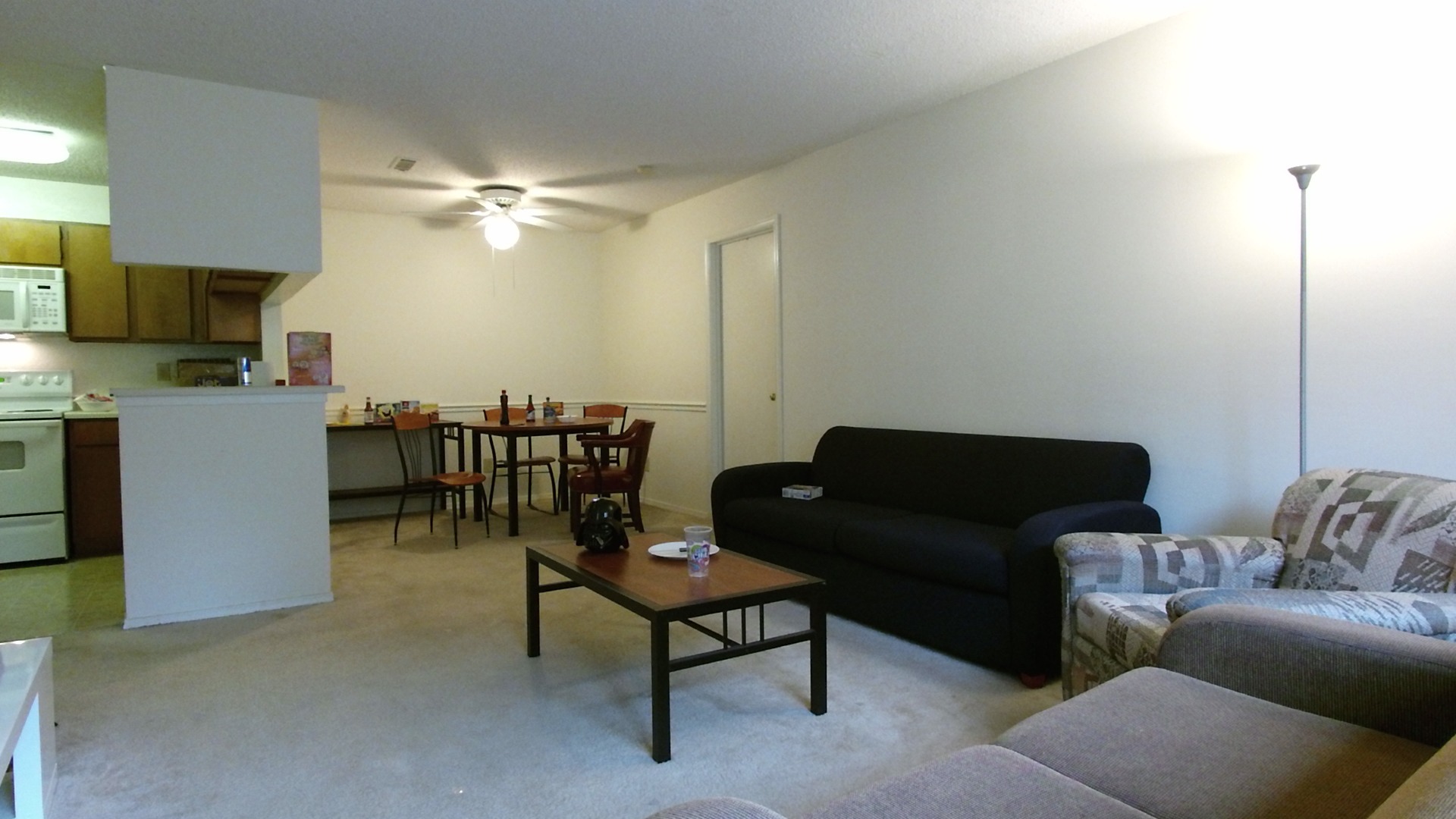}&
		\includegraphics[width=0.23\textwidth]{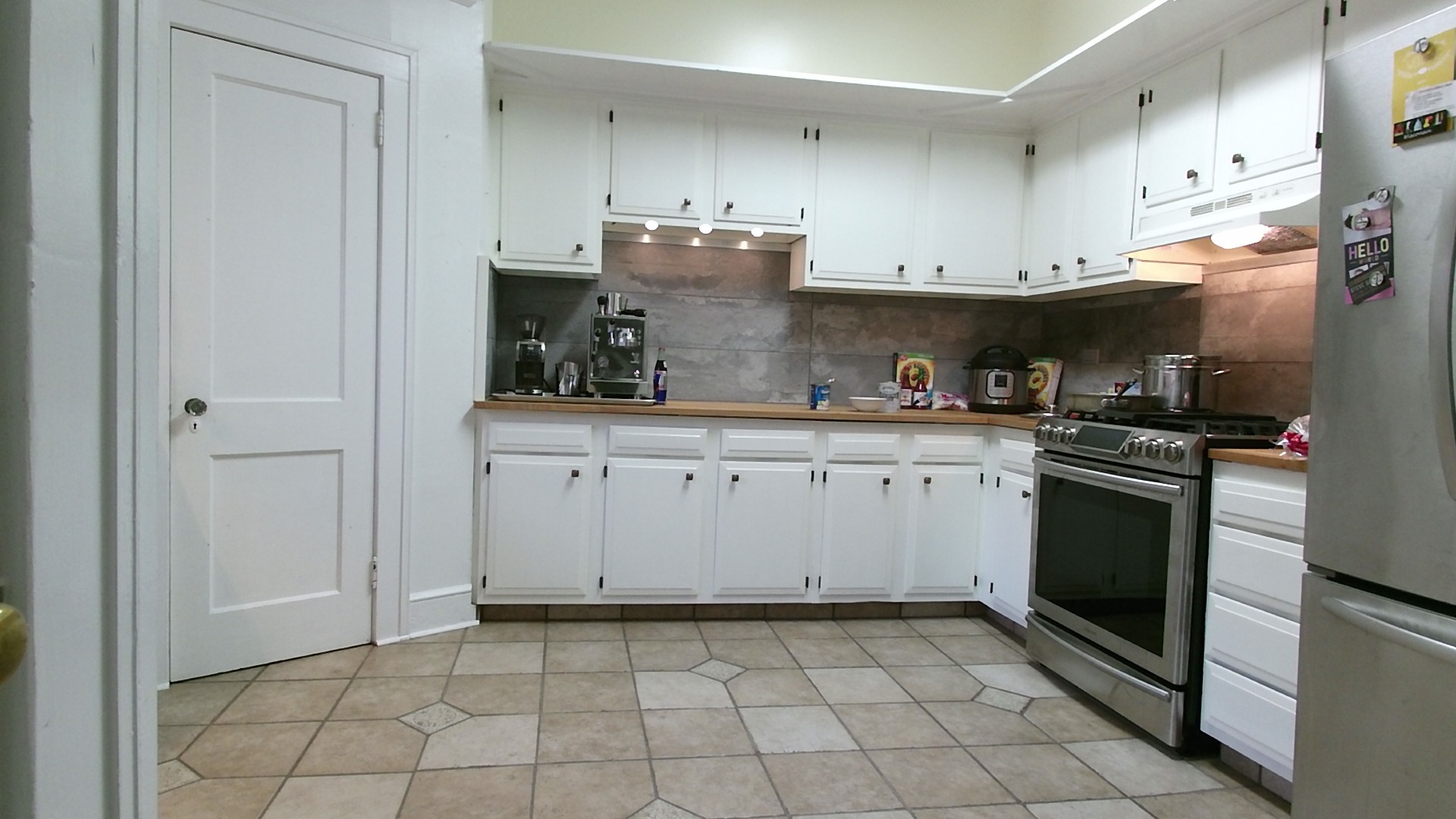}\\
		\includegraphics[width=0.23\textwidth]{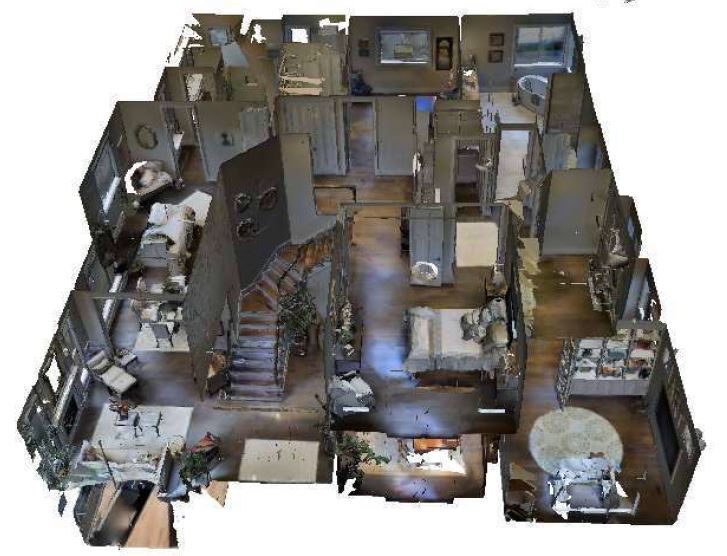}&
		\includegraphics[width=0.23\textwidth]{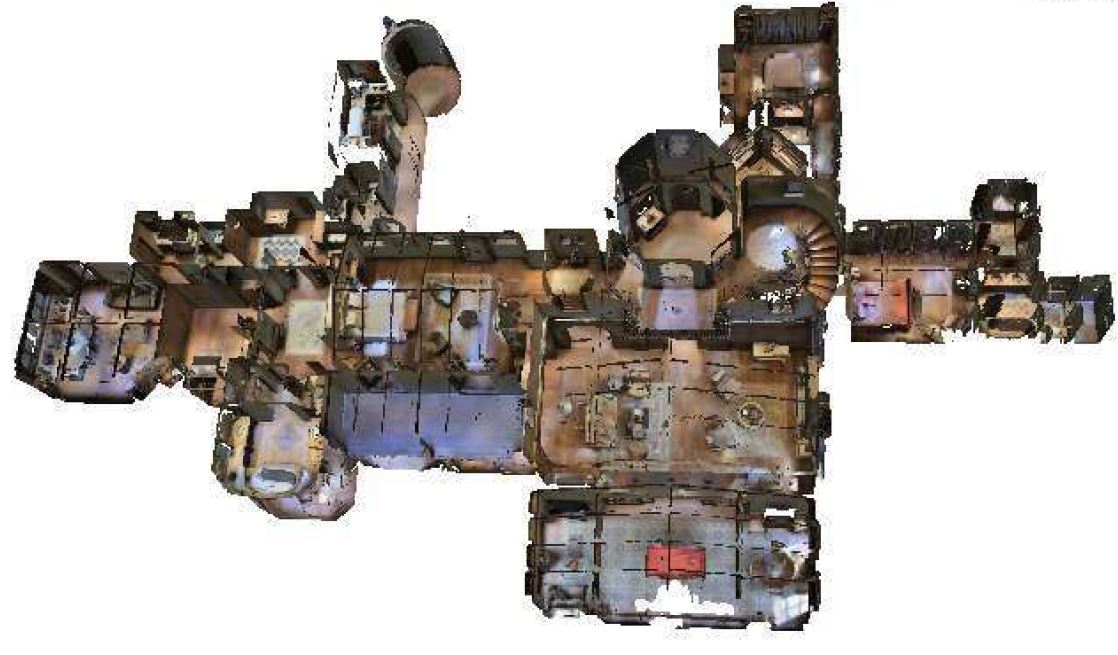}\\
   	\end{tabular}
\end{center}
\vspace{-0.3cm}
\caption{Evaluation dataset for AVS methods. The upper row shows sample images in {\em AVD}, captured with a real robotic platform in real apartments. The lower row shows sample 3D scans of complex scenes in {\em Habitat ObjectNav} Challenge.}
\vspace{-0.3cm}
\label{fig:dataset_samples}
\end{figure}

We evaluate our proposed \methname on two public datasets featuring real-world scenarios that have been commonly used for benchmarking methods addressing the AVS task. Active Vision Dataset (AVD)~\cite{ammirato2017dataset} is captured with a RGB-D sensor equipped on a robotic platform moving within real apartments in a grid manner, while Habitat ObjectNav~\cite{habitat19iccv} provides a simulator for rendering observations within 3D scans of real large indoor environments provided by Matterport3D dataset \cite{Matterport3D} (see Fig.~\ref{fig:dataset_samples}). 
We compare \methname against baselines and sota methods with both AVD (Section \ref{sec:exp:avd}) and Habitat ObjectNav (Section \ref{sec:exp:habitat}) under their corresponding evaluation protocol and performance metrics for fair comparison. Moreover, we conduct an ablation study (Section \ref{sec:exp:ablation}) where we show the impact of the proposed belief reinvigoration strategy and the impact of an imperfect object detector on AVS planning. 



\subsection{Evaluation on AVD}
\label{sec:exp:avd}
\begin{table*}[t]
\vspace{0.15cm}

\caption{Results on the three test scenes from AVD using the GT annotations for object detection. We report the results of EAT~\cite{schmid2019iros} and POMP \cite{wangpomp} as in their original paper. Results are averaged over multiple search episodes as defined in AVD benchmark, with the standard deviations presented within the parentheses. Results use known floor maps are highlighted in \textit{italic}. Best results of methods without known floor maps are highlighted in \textbf{bold}.}
\label{tab:AVD}
\resizebox{\textwidth}{!}{%
\begin{tabular}{|c|c|c|c|c|c|c|c|c|c|c|c|c|}
\hline
\multirow{2}{*}{} & \multicolumn{12}{c|}{AVD}  \\ \cline{2-13} 

\multirow{2}{*}{Method} & \multicolumn{3}{c|}{Easy} & \multicolumn{3}{c|}{Medium} & \multicolumn{3}{c|}{Hard} & \multicolumn{3}{c|}{Avg.} \\ \cline{2-13} 
 & SR $\uparrow$ & APL$\downarrow$ & ASPPL$\uparrow$  & SR$\uparrow$ & APL$\downarrow$ & ASPPL$\uparrow$ & SR$\uparrow$ & APL$\downarrow$ & ASPPL$\uparrow$ & SR$\uparrow$ & APL$\downarrow$ & ASPPL$\uparrow$ \\ \hline
\textit{POMP $(known\ env)$} \cite{wangpomp} & \textit{0.98} & \textit{13.6} & \textit{0.72 (0.26)} & \textit{0.73} & \textit{17.1} & \textit{0.8 (0.23)} & \textit{0.56} & \textit{20.5} & \textit{0.72 (0.39)} & \textit{0.76} & \textit{17.1} & \textit{0.75 (0.29)} \\ \hline\hline
Random Walk & 0.32 & 74 & 0.19 (0.35) & 0.11 & 74.48  & 0.21 (0.36)  & 0.10 & 79.27 & 0.17 (0.18) & 0.18 & 75.91 & 0.19 (0.29)  \\ \hline
EAT~\cite{schmid2019iros} & 0.77 & \textbf{12.2} & - & 0.73 & \textbf{16.2} & - & 0.58 & \textbf{22.1} & - & 0.69 & \textbf{16.8} & - \\ \hline
\textbf{\methname} & \textbf{1.0} & 14.27 & \textbf{0.70 (0.28)} & \textbf{0.79} & 19.4 & \textbf{0.76 (0.26)} & \textbf{0.84} & 29.54 & \textbf{0.61 (0.33)} & \textbf{0.87} & 21.07 & \textbf{0.68 (0.29)} \\ 
\hline
\end{tabular}%
}
\vspace{-0.2cm}

\end{table*}
AVD is the largest real-world dataset available for testing AVS, containing 17 scans of 9 real apartments recorded by a robot equipped with a RGB-D camera. For each scene, AVD provides the RGB-D images captured from all possible robot's poses, as well as the camera intrinsics and extrinsics. The action space $A$ in AVD is $\{$\texttt{Forward}, \texttt{Backward}, \texttt{Turn\_Left}, \texttt{Turn\_Right}, \texttt{Stop}$\}$, with a translation step of 30 cm and a rotation step of $30^{\circ}$.

\paragraph*{\textbf{Performance Metrics}}
In the AVD Benchmark (AVDB), a run is considered successful when the robot terminates its exploration in one of the ending positions specified in AVDB, which are closest to the object with the object appearing in the camera view. There are three main metrics defined for benchmarking different methods: \textbf{Success Rate (SR)}, is the ratio of successful episodes over the total number of episodes; \textbf{Average Path Length (APL)} is the average number of poses visited by the robot among the paths that lead to a successful search (a lower value indicates a higher efficiency); and \textbf{Average Shortest Predicted Path Length (ASPPL)} is the average ratio between the length of the shortest possible path to reach a valid destination pose provided by AVDB and the length of the path generated by the method (a larger value indicates a higher absolute efficiency).
We also compute the standard deviation of ASPPL to investigate the variability of path efficiency.

\paragraph*{\textbf{Baselines and Comparisons}}
We compare \methname against:
\begin{itemize}
\setlength\itemsep{0em}
\item \textbf{Random Walk}: a standard baseline where the robot move randomly for a certain number of steps.
\item \textbf{EAT} \cite{schmid2019iros}: a reinforcement learning approach trained on the remaining scenes of AVD. A Region proposal network (RPN) extracting object proposals is combined with a DRQN for the policy learning. 
\item \textbf{POMP} \cite{wangpomp}: a POMCP-based solver that only works with known environments. This method requires as input the 2D floor map and the reachable robot poses.
\end{itemize}

Note that among the RL-based strategies, we consider EAT~\cite{schmid2019iros} and GAPLE~\cite{ye2019ral} that both use AVD for evaluation (discussed in Sec.~\ref{sec:soa}). However, the evaluation protocol adopted by GAPLE is not documented, while EAT~\cite{schmid2019iros} shared their protocol explicitly. For a fair comparison, we follow the evaluation protocol in both EAT \cite{schmid2019iros} and POMP \cite{wangpomp}, which perform search on a subset of objects under three test scenes with an increasing difficulty level. The test scenes are: $i)$ Home\_005\_1, the easy scenario, consists only of a kitchen; $ii)$ Home\_001\_2, the medium scenario, the robot has to navigate a living room and an open kitchen; $iii)$ Home\_003\_2 is the hard scenario, where the robot has to explore a living room, a half-open kitchen, a dining room, hallway and bathroom.
As in\cite{schmid2019iros} and \cite{wangpomp}, to only compare the goodness of the motion policy without the nuisances caused by underlying object detectors, we use the ground-truth annotations of the object for the evaluation, and we limit the search to 125 steps. 
For POMP++ we set the reward value to +1000 for when the object is detected and -1 for when it is not. 
The impact of a real-world detector on AVD is the subject of another experiment in Section \ref{sec:exp:ablation}. 

\paragraph*{\textbf{Results Discussion}}
Table~\ref{tab:AVD} shows the performances achieved by all methods on the three test scenes in AVD.
We observe that on average \methname outperforms all the other methods in term of Success Rate. Even in complex episodes, where the target object is initialised far away from the robot, \methname is able to succeed the search task (see Supplementary Material for more search dynamics). Our path efficiency on the successful searches is lower compared to other methods as \methname has to explore more areas to look for the object, while POMP \cite{wangpomp} exploits a known 2D floor map and EAT has already encoded the environment knowledge via its offline training. 
%
The improvement of \methname in terms of SR over POMP is mainly contributed by our new belief reinvigoration strategy. We provide more details in Section~\ref{sec:exp:ablation}.

\subsection{Evaluation on Habitat ObjectNav}
\label{sec:exp:habitat}
Habitat~\cite{habitat19iccv} is a simulation platform designed for embodied AI, where robots can move and observe within realistic 3D environments, \eg 3D scans of real large scenes from Matterport3D~\cite{Matterport3D} and Gibson~\cite{xiazamirhe2018gibsonenv}. In particular, the Habitat ObjectNav Challenge makes use of 11 scans of complex scenes in Matterport3D for validating methods (see the second row in Fig. \ref{fig:dataset_samples}). 
While the robot moves in the Habitat environment, the simulator renders the RGB-D images at each step taken from the current pose, as well as the camera intrinsics and extrinsics. The action space $A$ in Habitat ObjectNav Challenge is the set of four actions $\{$\texttt{Forward}, \texttt{Turn\_Left}, \texttt{Turn\_Right}, \texttt{Stop}$\}$. 
We set the translation step to $25$ cm and the rotation step to $30^{\circ}$.

\paragraph*{\textbf{Performance Metrics}}
In the Habitat ObjectNav Challenge, a success is defined if the distance between the robot and the target object is $\leq1$m, and the object is within the camera’s field of view, without specifying the ending poses as in AVD. There are three main performance metrics defined for the Habitat ObjectNav Challenge:
\textbf{Success Rate (SR)}, is the ratio of successful runs over the total number of runs, the same as in AVD metrics; \textbf{Success weighted Path Length (SPL)}, measures the path efficiency of successful episodes\footnote{SPL is used as the main criterion in the Habitat ObjectNav Challenge}, defined as $SPL = \frac{1}{N}\sum_{i=1}^{N}S_i\frac{l_i}{\max(l_i^r,l_i)}$,
where $l_i$ is the length of the shortest path between the goal and the target for an episode, $l_i^r$ is the length of path taken by an robot in an episode, and $S_i$ is the binary indicator of a success in episode $i$. 
Finally, \textbf{Distance To Success (DTS)} is further defined by~\cite{chaplot2020object} which measures the distance of the robot from the success threshold boundary when the episode ends. DTS is computed as $\max(\left \|x_{T}-x_{G}\right\|-d_{s},0)$, where $x_{T}$ is the robot location at the termination, $x_{G}$ is the goal location at the end of the episode, and $d_s$ is the success threshold.


 

\paragraph*{\textbf{Baselines and comparisons}}
We compare \methname against the winner entry in the Habitat ObjectNav Challenge, \textbf{SemExp}~\cite{chaplot2020object}, as well as baselines reported in the literature:
\begin{itemize}
\setlength\itemsep{0em}
    \item \textbf{Random walk}: standard baseline where the robot moves randomly for a certain number of steps.
    \item \textbf{RGBD + RL}~\cite{habitat19iccv}: a vanilla recurrent RL policy initialised with ResNet18 backbone followed by a GRU.
    \item \textbf{RGBD+Semantics+RL}: an adaptation from~\cite{rgbd-sem-rl} which passes semantic segmentation and object detections along with RGBD input to a recurrent RL policy. 
    \item \textbf{Classical Mapping + FBE}~\cite{frontier1997}: classical robotics pipeline for mapping followed by frontier-based exploration (FBE). When the detector finds the object, a local policy reaches the object with an analytical planner.
    \item \textbf{Active Neural SLAM}~\cite{chaplot2020learning}: an exploration policy trained to maximise the coverage. Whenever the target object is detected, the same local policy as described above is applied to approach the object.
    \item \textbf{SemExp}~\cite{chaplot2020object}: a RL policy learner with a semantic mapping of the explored environment, and embedded object-to-object priors.
\end{itemize}

We follow the evaluation protocol defined in~\cite{chaplot2020object}, where a subset of object categories that are common between MP3D and MS-COCO is chosen as target: $\{$\emph{chair}, \emph{couch}, \emph{potted plant}, \emph{bed}, \emph{toilet}, \emph{tv}$\}$. For object detection, we used Mask-RCNN with ResNet50 backbone pretrained on MS-COCO. Experiments are performed with 11 scenes in the validation set of Matterport3D, following the configuration in terms of robot starting pose and target object, provided by the Habitat ObjectNav Challenge. The walls are reconstructed online with the ground-truth semantic segmentation. The reward value is set to +1000 for when the object is detected and -1 for when it is not.

\begin{table}[t]
\centering
\vspace{0.15cm}

\caption{Results of \methname and baselines evaluated on 11 test scenes of Matterport3D, with the object detector pretrained on MS-COCO.}
\label{tab:habitat}
\begin{tabular}{|l|c|c|c|}
\hline
&  \multicolumn{3}{c|}{Habitat}\\
\cline{2-4} Method &  SPL $\uparrow$ & SR $\uparrow$ & DTS (m)$\downarrow$ \\ \hline
Random &  0.005 & 0.005  & 8.048  \\\hline
RGBD + RL~\cite{habitat19iccv}  &  0.017  & 0.037 & 7.654 \\\hline
RGBD + Sem + RL\cite{rgbd-sem-rl} &  0.015 & 0.031 & 7.612\\\hline
Classical Mapping + FBE~\cite{frontier1997} &  0.117 & 0.311 & 7.102 \\\hline
Active Neural SLAM\cite{chaplot2020learning}    &  0.119  & 0.321  & 7.056  \\\hline
SemExp \cite{chaplot2020object} &  0.144 & 0.360 & 6.733 \\\hline\hline
\textbf{\methname} &  \textbf{0.148} & \textbf{0.420} & \textbf{3.726} \\\hline
\end{tabular}
\end{table}

\paragraph*{\textbf{Result Discussion}}
Table \ref{tab:habitat} shows the results evaluated on the 11 test scenes in Habitat ObjectNav dataset.
Our approach has a SR 6\% higher than the best RL approach SemExp~\cite{chaplot2020object} and 10\% higher than the Active Neural SLAM~\cite{chaplot2020learning}. In terms of SPL, we are just slightly better than SemExp, which indicates that our path efficiency could be lower given the higher SR we achieve. 
Yet, our Distance To Success (DTS) is notably lower than all the other approaches, meaning that our failed episodes tend to terminate nearer to the target location.

It is noticeable that the overall SR achieved by all methods are low due to the unsatisfactory 3D reconstruction quality of scenes (\eg incomplete scans) and sometimes confusing object annotations (\eg "bulletin board" as "tv\_monitor"). The use of a standard object detector without fine-tuning on the rendered images also contributes to the low performance since the detector is prone to errors. The impact of the object detector on AVS task is studied in Section~\ref{sec:exp:ablation}. 

Moreover, we notice that algorithms that use explicit semantic mapping of the environment, such as Classical Mapping+FBE \cite{frontier1997}, Active Neural SLAM \cite{chaplot2020learning}, SemExp \cite{chaplot2020object} and \methname are able to perform significantly better than RL-based methods \cite{habitat19iccv} that rely on implicit scene representation via visual features extracted from the sequence of observations. 
However, as the scenes in Habitat ObjectNav varies a lot among each other, in terms of the scene types and dimensions, covering modern/ancient homes, palaces and even a spa. The merits of encoding sophisticated scene semantics can be limited, which allows \methname to outperform other methods with only 2D wall maps that are reconstructed online.
Please refer to Supplementary Material for the state dynamics of both successful and failed searches as robot moves within the test scenes.


\subsection{Ablation study}
\label{sec:exp:ablation}

\begin{table}[t]
\centering
\vspace{0.15cm}
\caption{Results of \methname and POMP+ on the Hard scenario from AVD, with both GT annotations and an object detector \cite{ammirato2017dataset}.}
\label{tab:ablation}
\begin{tabular}{|l|c|c|c|}
\hline
&  \multicolumn{3}{c|}{AVD (Hard)}\\
\cline{2-4} Method &  SR $\uparrow$ & APL $\downarrow$  & ASPPL $\uparrow$ \\ \hline
POMP (GT, $known\ env$) & 0.56 & 20.05 & 0.72(0.39)\\\hline 
POMP+ (GT, $known\ env$) &  0.86 & 33.71 & 0.55(0.35)\\\hline
\methname (GT) & 0.84 & 29.54 & 0.61(0.33)\\\hline\hline
POMP (Detector, $known\ env$) &0.37 & 22.01& 0.63(0.29)\\\hline
POMP+(Detector, $known\ env$) &  0.45 & 23.95 & 0.63(0.41)\\\hline
\methname (Detector) &  0.41 & 22.41 & 0.63(0.31)\\\hline
\end{tabular}
\vspace{-0.5cm}
\end{table}

We mainly evaluate the effectiveness of the proposed belief reinvigoration strategy and the impact of imperfect object detector. We compare \methname to POMP \cite{wangpomp} and POMP+, which is POMP for known environments with our novel belief reinvigoration strategy, with both ground-truth annotation and the object detector provided by AVD \cite{ammirato2017dataset}. 
Table \ref{tab:ablation} reports the averaged results obtained with the hard scene Home\_003\_2 in AVD. POMP+ with the novel belief reinvigoration strategy improves the SR by 30\% although at the cost of lengthier paths. SR is significantly improved by the new reinvigoration strategy because the original one tends to vanish particles that are not covered by successful simulations, which limits the potential of a successful search, as we explained in Section \ref{sec:method:belief_resampling}. \methname achieves a slightly worse SR compared to POMP+ due to the lack of a known 2D floor map.

Moreover, the performance of the detector imposes a direct impact on the AVS performance, where we can observe the SR is reduced almost by half compared to the results obtained with GT annotations. This is because during the POMCP exploration, the output of the detector is used to either prune candidate object location in case of no detection; or as a stopping condition for the exploration if the target object is detected.


\subsection{Real-time performance analysis} 
The complexity of \methname is $O(\alpha\times(\beta\times \mu +\delta))$, where $\alpha$ is the number of steps performed in the real environment, $\beta$ is the number of simulations performed for each step, $\mu$ is the max number of steps per simulation, $\delta$ is the number of actions to update the state space, the map, the transition and observation models. 
Note that all parameters can be tuned to satisfy real time performance when scaling the algorithm to large environments.
When experimenting on AVD that contains realistic indoor housing environments, \eg typical homes, we achieve 0.07 seconds per step on average with a standard machine with a 6-cores Intel i7-6800k CPU. Such processing speed is considered sufficient for real-time robotic applications. 


\section{Conclusion}
\label{sec:conclusion}
In this work we proposed a POMCP-based planner, \methname, that solves the AVS problem in unknown environments, without the need of offline training. We introduced novel modifications to the POMDP's formulation, allowing for the search space to be dynamic, and grow as the robot explores the environment. Following this new formulation, we proposed a novel way to update the belief in the underlying POMDP solver, POMCP, which boosts the correct coverage of the expanding search space. We outperformed the sota methods on two benchmark datasets in terms of the success rate by a significant margin. As future work, we will further integrate scene graphs within the POMCP framework to better inject scene semantic for boosting the search efficiency of our method. 
\\

{\small
\bibliographystyle{IEEEtran}
\bibliography{refs_short}
}

\end{document}